\documentclass[runningheads]{llncs}
\usepackage[T1]{fontenc}

\usepackage{graphicx,verbatim,amsmath}

\usepackage{pifont}         
\usepackage{enumitem}

\newcommand{\cmark}{\ding{51}}  
\newcommand{\nmark}{--}         
\usepackage{hyperref}
\usepackage[table]{xcolor}
\usepackage{booktabs}
\usepackage{multirow}
\usepackage{graphicx}
\usepackage{dsfont}
\usepackage{amssymb}
\usepackage{amsmath} 
\usepackage{tikz} 
\definecolor{BaseTeal}{HTML}{64B4AA}

\begin{document}

\title{Displacement Preserving Relational Distillation for Robust Medical Segmentation}
\titlerunning{DPRD for Robust Medical Segmentation}

\makeatletter
\renewcommand{\@fnsymbol}[1]{%
  \ifcase#1
  \or *
  \or \ensuremath{\dagger}
  \or \ensuremath{\ddagger}
  \else\@ctrerr\fi}
\makeatother

\author{Zhicheng Ding\inst{1,3}\thanks{Equal contribution.} \and
Xinyu Chu\inst{1,3}\textsuperscript{*}\and
Jung Im Choi\inst{2}\and
Qing Tian\inst{3}\textsuperscript{†}\and
Tianyu Shi\inst{4}\and
Xiaoqian Jiang\inst{6}\and
Lijing Zhu\inst{5}\and
Qizhen Lan\inst{6}\textsuperscript{*}\thanks{Corresponding authors.}}


\authorrunning{Z. Ding et al.}

\institute{
Bowling Green State University, Bowling Green, OH, USA\\
\and
The University of Findlay, Findlay, OH, USA\\
\and
University of Alabama at Birmingham, Birmingham, AL, USA\\
\email{qtian@uab.edu}\\
\and
University of Toronto, Toronto, Ontario, Canada\\
\and
University of Houston - Clear Lake, Houston, TX, USA\\
\and
UTHealth Houston, Houston, TX, USA\\
\email{qizhen.lan@uth.tmc.edu}}

\maketitle

\begin{abstract}


Accurate 3D medical segmentation is limited by anatomical variability and high computational costs. While knowledge distillation (KD) offers a route for model compression, conventional methods often fail to preserve complex structures and are overwhelmed by background noise. We propose Displacement-Preserving Relational Distillation (DPRD), which distills latent anatomical trajectories via vector based alignment to preserve the orientation and relative scale of the teacher’s manifold, and prevents signal dilution by anchoring distillation in task-relevant structures. Integrated into nnU-Net, DPRD outperforms established baselines on ISLES 2022 and AMOS 2022 benchmarks. Notably, on the AMOS dataset, DPRD achieves a Dice score of 85.46\%, edging out the high-capacity MedNeXt teacher while significantly reducing boundary errors. Despite utilizing only $\sim$5\% of the teacher’s parameters and $\sim$3\% of its FLOPs, our approach maintains high structural consistency. This provides a robust, efficient solution for deploying high performance segmenters in resource-constrained clinical environments. Code: \url{https://github.com/ClinicaAlpha/DPRD-3D-MedSeg}

\keywords{3D Medical Image Segmentation \and Knowledge Distillation}

\end{abstract}

\section{Introduction}

Clinicians must precisely localize 3D operative targets and at-risk structures during image-guided procedures, often under strict workflow constraints \cite{lan2026performance,mezger2013navigation,nitsch2019mrius}. This necessitates reliable 3D segmentation methods robust to heterogeneous imaging and efficient for routine deployment \cite{guan2021domain}. While fundamental to 3D quantitative imaging tasks like organ volumetry and therapy planning \cite{litjens2017survey,milletari2016vnet,isensee2021nnunet}, challenges from domain shift and the high computational footprint of modern 3D backbones hinder resource-efficient deployment \cite{guan2021domain,hatamizadeh2022swinunetr,lee2022uxnet,roy2023mednext}.

Knowledge distillation (KD) provides a principled route to mitigate the computational burden of 3D models by transferring supervision from high-performing teachers to efficient student segmenters \cite{hinton2015distilling,qin2021_tmi_kdseg,noothout2022_jmi_kdensembles,li2026comprehensive}.
However, KD for dense 3D segmentation faces distinctive obstacles. First, voxel-wise distillation (e.g., logit or feature alignment) assumes that teacher and student representations are structurally compatible. Under architectural heterogeneity, differing feature scales or semantics can render strict activation matching unstable \cite{liu2019skdseg,he2019_kadseg,shu2021cwd}. 
Second, 3D segmentation is often severely class-imbalanced: most voxels belong to background, so uniform alignment or naive pooling can dilute supervision, especially for small targets \cite{sudre2017_gdl}. Third, medical imaging data are subject to cross-scanner appearance drift. In this setting, absolute activation imitation may overfit to style-specific cues and acquisition idiosyncrasies rather than preserving transferable, clinically meaningful structures \cite{guan2021domain,dou2020_ummkd,jia2026unsupervised}.

These limitations suggest that effective distillation for 3D segmentation should rely less on absolute, voxel-aligned activation imitation and more on transferable structure. Beyond producing accurate per-case outputs, a strong teacher often induces a representation space where clinically similar cases form coherent neighborhoods \cite{tung2019spkd,zhu2025pathology}. Prior work has shown that distilling inter-sample relations (e.g., pairwise similarities or distances) can provide a softer, architecture-agnostic signal than pointwise feature matching \cite{park2019rkd,tung2019spkd,tian2020crd,liu2019skdseg,ding2026dfm,lan2025acam}. However, applying existing relational KD methods from natural imaging to 3D medical data presents significant bottlenecks. Conventional RKD \cite{park2019rkd} transfers structural knowledge by penalizing differences in scalar distances or angles, which collapses high-dimensional anatomical variations into isotropic scalars. Similarly, CIRKD \cite{yang2022cirkd} constructs dense pixel-level relations for 2D images; however, this becomes computationally prohibitive and inherently noisy in 3D segmentation, where volumes are overwhelmingly dominated by non-informative background voxels.

To address these gaps, we propose \emph{Displacement Preserving Relational Distillation (DPRD)} for robust 3D medical segmentation. DPRD 
performs vector displacement matching on ROI-masked case-level embeddings across hierarchical encoder stages. By pooling encoder features into task-relevant ROI embeddings and aligning all-pairs batch-wise displacements, DPRD avoids direct activation matching and scalar-only relational supervision while reducing background dilution. Coupling trajectory alignment with batch-wise scale normalization provides scale-invariant local relational supervision. Ablations support the contributions of trajectory modeling and ROI masking. Our contributions are:

\begin{itemize}[noitemsep, topsep=0pt, parsep=0pt, partopsep=0pt, leftmargin=*]
  \item We introduce \emph{displacement-preserving} relational distillation for 3D segmentation, using batch-wise displacement vectors with relational scale normalization to impose a scale-invariant local relational constraint rather than matching absolute activations.
  \item We develop a practical cross architecture, ROI-aware, multi-stage distillation design that is memory efficient for 3D volumes and explicitly mitigates background dominated supervision.
  \item We empirically show consistent improvements across ISLES and AMOS benchmarks and provide ablation analysis on the contributions of ROI-aware masking, displacement alignment, and distance consistency.
\end{itemize}

\section{Method}
\begin{figure}[t]
\centering
\includegraphics[width=0.95\textwidth]{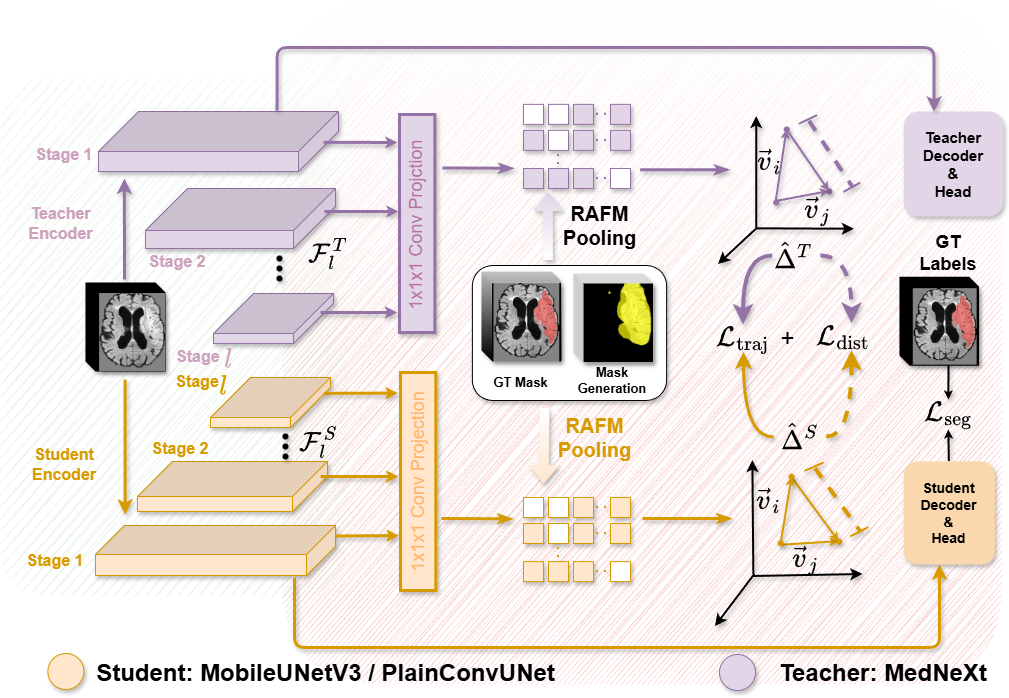}
\caption{Overview of DPRD. RAFM obtains ROI-aware case-level embeddings, and DPRA aligns pairwise displacement relations between teacher and student representations.} \label{figdprd}
\end{figure}

We propose Displacement-Preserving Relational Distillation (DPRD) for high performance, lightweight 3D medical segmentation (Fig.~\ref{figdprd}). Unlike restrictive voxel-wise alignment, DPRD matches batch-wise displacement vectors after relational scale normalization in a common projected embedding space, providing scale-invariant local relational supervision between teacher and student representations. To handle architectural and scanner variability, we introduce: (1) \emph{Displacement Preserving Relational Alignment}, which matches teacher-induced local displacement relations instead of absolute activations; and (2) \emph{ROI-Aware Feature Masking}, which uses anatomical priors to focus alignment on task-relevant structures and mitigate background dilution. These modules ensure multi-stage structural consistency, enabling the student to learn 3D anatomical variances.

\subsection{Displacement-Preserving Relational Alignment (DPRA)}

\subsubsection{Pairwise Displacement Mapping}

To facilitate architecture-agnostic knowledge transfer, we first extract a set of feature embeddings $\{\mathbf{v}_i^\phi\}_{i=1}^N$ ($\phi \in \{T, S\}$) (Detail in Sec 2.2) from the $l$-th stage of the teacher and student encoders. Unlike conventional relational knowledge distillation that collapses complex sample relationships into scalar distances or angles, our method models local relational cues by calculating pairwise displacement vectors between ROI-pooled case-level embeddings for all unique pairs $(i, j)$:
\begin{equation}
\Delta_{ij}^\phi = \mathbf{v}_i^\phi - \mathbf{v}_j^\phi, \quad \phi \in \{T, S\}
\end{equation}
\noindent where $1 \le i < j \le N$, for a mini-batch of size $N$, this yields $K = N(N-1)/2$ displacement vectors. In the teacher's latent space, $\Delta_{ij}^T$ represents a high-dimensional relational bridge that captures structural patterns less dependent on specific convolutional operations. The student’s task is to reconstruct these ``anatomical trajectories'' within its own latent space, ensuring that the directional evolution from one case to another remains consistent across different feature hierarchies.

\subsubsection{Heterogeneous Batch Normalization}

To ensure stable convergence when distilling knowledge between architectures with different activation scales, we introduce a batch-wise normalization scheme to eliminate intrinsic activation bias. For each stream $\phi$, we calculate the batch-level relational scale $\mu^\phi$ as the average $L_2$ norm across all $K$ displacement vectors within the current batch:
\begin{equation}
\mu^\phi = \frac{1}{K} \sum_{1 \le i < j \le N} \|\Delta_{ij}^\phi\|_2
\end{equation}
The normalized displacement vectors are then defined as: 
\begin{equation}
\hat{\Delta}_{ij}^\phi = \Delta_{ij}^\phi / (\mu^\phi + \epsilon)
\end{equation}
This normalization removes global activation-scale differences, encouraging the student to focus on scale-invariant local displacement relations rather than absolute feature magnitudes. By aligning the scaled displacements, we encourage the distillation process to exhibit enhanced stability against intensity fluctuations and scale discrepancies common in multi-vendor medical imaging data.

\subsubsection{Relational Distillation Objective}

To provide scale-invariant local relational supervision, the relational loss $\mathcal{L}_{rel}^l$ at stage $l$ is composed of two synergistic components:

\begin{itemize}
    \item[(1)] \textbf{Pairwise Displacement Loss ($\mathcal{L}_{traj}$):} This term functions as a fine-grained trajectory reconstruction, constraining the student to align the precise orientation and normalized length distribution of the teacher's displacement vectors across all latent dimensions. We compute the cumulative error across the shared channel dimension $C$ using the outlier-robust Smooth L1 loss $\Psi$:
    \begin{equation}
    \mathcal{L}_{traj} = \frac{1}{K} \sum_{1 \le i < j \le N} \left( \sum_{c=1}^{C} \Psi(\hat{\Delta}_{ij,c}^S - \hat{\Delta}_{ij,c}^T) \right)
    \end{equation}

    \item[(2)] \textbf{Distance Consistency Regularization ($\mathcal{L}_{dist}$):} While the aforementioned normalization reduces overall scaling differences, $\mathcal{L}_{dist}$ explicitly preserves the relative length structure within the batch by aligning the relative Euclidean distances of the displacements. This provides a complementary constraint on the relational scale:
    \begin{equation}
    \mathcal{L}_{dist} = \frac{1}{K} \sum_{1 \le i < j \le N} \Psi(\|\hat{\Delta}_{ij}^S\|_2 - \|\hat{\Delta}_{ij}^T\|_2)
    \end{equation}
\end{itemize}

The total stage-wise relational loss is defined as $\mathcal{L}_{rel}^l = \lambda_{traj} \mathcal{L}_{traj} + \lambda_{dist} \mathcal{L}_{dist}$, where $\lambda_{traj}$ and $\lambda_{dist}$ are balancing hyperparameters. This composite objective encourages the student to match teacher-induced local displacement directions and relative distance patterns without requiring global manifold preservation.

\subsection{ROI-Aware Feature Masking (RAFM)}
\subsubsection{Density-Normalized Feature Aggregation}
To bridge the architectural discrepancy between the teacher and student, we project their feature maps into a shared 128-channel space via $1 \times 1 \times 1$ convolutions, yielding channel-aligned maps $\mathbf{\hat{F}}^T$ and $\mathbf{\hat{F}}^S$. We then employ a density-normalization scheme to produce size-invariant embeddings from these projected maps. This ensures that the subsequent displacement vectors represent inter-case anatomical variance rather than differences in voxel count or occupied volume.
\begin{equation}
\mathbf{v}^\phi = \frac{\mathcal{P}(\mathbf{\hat{\mathbf{F}}}^\phi \odot \mathbf{M}^\phi)}{\mathcal{P}(\mathbf{M}^\phi) + \epsilon}
\end{equation}
where $\mathcal{P}(\cdot)$ denotes the pooling operation used to obtain compact case-level embeddings, and $\epsilon$ is a small constant for numerical stability. By reweighting the pooled features by the mask density $\mathcal{P}(\mathbf{M}^\phi)$, this strategy yields a compact, scale-invariant representation that encapsulates the average activation intensity within the anatomical structures, providing a stable foundation for relational alignment.

\subsubsection{Anatomical Mask Construction}
To ground distillation in relevant anatomy, we derive a unified binary mask $\mathbf{M}$ by merging all foreground ground-truth categories:
\begin{equation}
\mathbf{M}_{raw} = \mathds{1} \left( \sum_{c=1}^{C_{class}} \mathbf{Y}_c > 0 \right)
\end{equation}
where $\mathbf{Y}$ represents the segmentation target. To align with the respective spatial resolutions of the teacher and student feature maps, $\mathbf{M}_{raw}$ is rescaled using nearest-neighbor interpolation to generate model specific masks: $\mathbf{M}^T$ and $\mathbf{M}^S$. This localization ensures that the anatomical ROI is precisely captured within each model's unique coordinate system before the aggregation process. It is important to note that the ROI-aware masking is exclusively utilized during the training phase to guide the relational alignment.

\subsection{Multi-Stage Joint Optimization}

To capture hierarchical anatomical structures, we apply DPRD across all stages in the encoder set $L$. We employ stage-specific coefficients to balance the relational signals across varying feature resolutions. The student model is optimized by minimizing a composite loss function:

\begin{equation}
\mathcal{L}_{total} = \mathcal{L}_{seg} + \sum_{l \in L} \omega^l \left( \lambda_{traj} \mathcal{L}_{traj}^l + \lambda_{dist} \mathcal{L}_{dist}^l \right)
\end{equation}

where $\mathcal{L}_{seg}$ denotes the standard nnU-Net objective (Dice and Cross-Entropy). Stage-wise weights $\omega^l$ for $l \in \{0, \dots, 5\}$ are set to $\{0.05, 0.1, 0.1, 0.15, 0.25, 0.35\}$. We assign higher weights to deeper layers, which typically encode more abstract anatomical patterns. Balancing weights are fixed at $\lambda_{traj}=1.0$ and $\lambda_{dist}=0.5$. This multi-scale consistency ensures the student prioritizes anatomical variances robust to acquisition noise and heterogeneous multi-center benchmarks.

\section{Experiment and Results}
\noindent \textbf{Datasets and Metrics} 

We evaluate DPRD on ISLES 2022 \cite{petzsche2022isles} (200 training/50 validation cases for MRI stroke lesion segmentation) and AMOS 2022 \cite{ji2022amos} (240 training/60 validation cases for 15-class abdominal organ segmentation). Performance is assessed via Dice Similarity Coefficient (DSC), 95\% Hausdorff Distance (HD95), and Normalized Surface Distance (NSD).

\noindent \textbf{Implementation Details.}

All experiments are implemented in nnU-Net~\cite{isensee2021nnunet}. For cross-architecture distillation, we use a pre-trained, frozen MedNeXt as the teacher and evaluate two heterogeneous students: MobileUNetV3 and PlainConvUNet. Following the nnU-Net validation protocol, results are reported on the standard data partitioning. Batch size is set to \textbf{4} for MobileUNetV3 and \textbf{2} for PlainConvUNet due to heterogeneous 3D memory footprints. DPRD provides stochastic local relational supervision, where mini-batch displacement constraints are accumulated across sampled cases, augmented patches, encoder stages, iterations, and epochs. The $O(N^2)$ relational alignment is lightweight because it operates on compact case-level embeddings rather than dense voxel-wise feature maps. During inference, the teacher and projection layers are removed, so the deployed model has the same parameters and FLOPs as the corresponding student. All models are trained with SGD. We compare DPRD with Logits KD~\cite{hinton2015distilling}, FitNet~\cite{romero2015fitnetshintsdeepnets}, RKD~\cite{park2019rkd}, and CIRKD~\cite{yang2022cirkd}, implemented under the same nnU-Net training setting with hyperparameters adapted for 3D medical volumes.

\begin{figure}[t]
\centering
\includegraphics[width=0.8\textwidth]{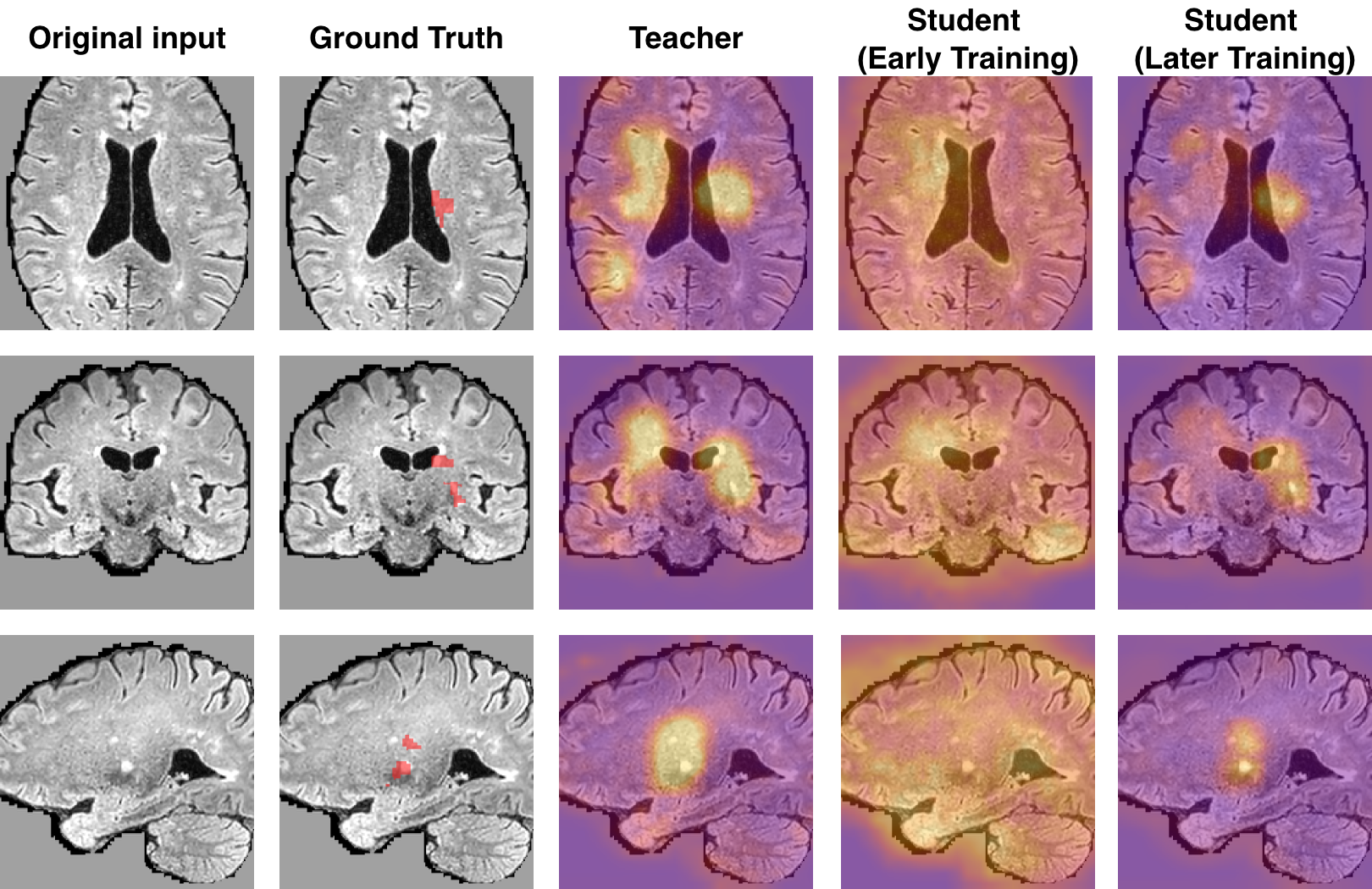}
\caption{Visualization of DPRD on ISLES 2022 at encoder stage 3 for a sample case in each orthogonal view (axial, coronal, sagittal). Columns: input MRI, ground truth (red), teacher, early-stage student, and late-stage student responses. Yellowish regions denote areas of higher importance; the late-stage student focuses on lesion regions more strongly than the early-stage student and aligns better with the ground truth.} \label{fig:isles_attention}
\end{figure}

\begin{figure*}[t]
\centering
\renewcommand{\arraystretch}{1.0}

\begin{tabular*}{\textwidth}{@{} @{\extracolsep{\fill}} cccccc @{}}
\parbox{0.16\textwidth}{\centering\fontsize{8}{10}\selectfont GT} &
\parbox{0.16\textwidth}{\centering\fontsize{8}{10}\selectfont Logits KD} &
\parbox{0.16\textwidth}{\centering\fontsize{8}{10}\selectfont FitNet} &
\parbox{0.16\textwidth}{\centering\fontsize{8}{10}\selectfont RKD} &
\parbox{0.16\textwidth}{\centering\fontsize{8}{10}\selectfont CIRKD} &
\parbox{0.16\textwidth}{\centering\fontsize{8}{10}\selectfont DPRD} \\[4pt]

\includegraphics[width=0.16\textwidth]{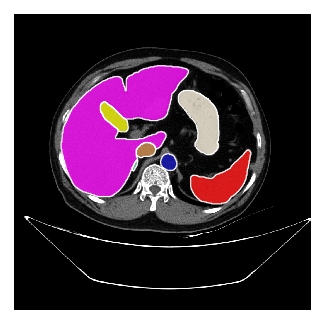} &
\includegraphics[width=0.16\textwidth]{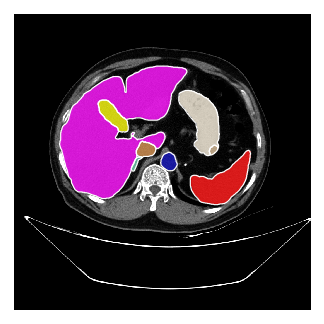} &
\includegraphics[width=0.16\textwidth]{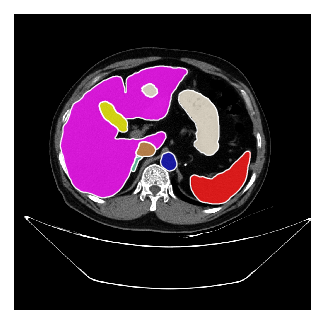} &
\includegraphics[width=0.16\textwidth]{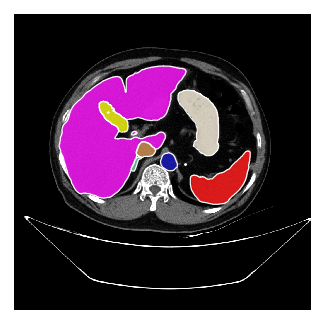} &
\includegraphics[width=0.16\textwidth]{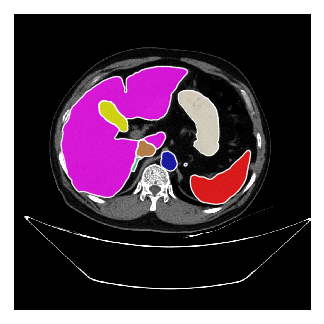} &
\includegraphics[width=0.16\textwidth]{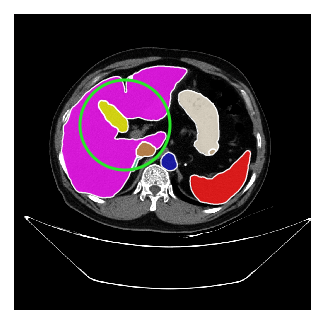} \\
\end{tabular*}

  \caption{Qualitative comparison on AMOS. Colors indicate different organs and are consistent across columns. Compared to other distillation baselines, our \textbf{DPRD} produces segmentation masks that are more consistent with the ground truth, particularly for the structures highlighted by the green circles (e.g., gallbladder and postcava).}
  \label{fig:amos_qual_case}
\end{figure*}

\subsection{Results}

\begin{table}[t]
\centering
\caption{Comparison of distillation methods on ISLES 2022.}
\label{tab:isles}

\begin{tabular}{l@{\hspace{-10pt}}c@{\hspace{-1pt}}c@{\hspace{-1pt}}ccc}
\toprule
Method & Params(M) & FLOPs(G) & Dice(\%) $\uparrow$ & NSD(2mm) $\uparrow$ & HD95(mm) $\downarrow$ \\ 
\midrule
Teacher-MedNeXt & 12.10 & 974.58 & 76.88$\pm$2.84 & 85.64$\pm$2.55 & 14.12$\pm$3.11 \\
Baseline-PlainConvUNet & 1.95 & 497.79 & 72.92$\pm$3.09 & 83.34$\pm$2.87 & 17.21$\pm$3.57 \\
\hline
Logits KD \cite{hinton2015distilling} & \multirow{5}{*}{1.95} & \multirow{5}{*}{497.79} & 72.02$\pm$3.22 & 82.61$\pm$3.12 & 17.49$\pm$3.74 \\
FitNet \cite{romero2015fitnetshintsdeepnets} & & & 73.93$\pm$3.05 & 84.29$\pm$2.82 & 16.79$\pm$3.57 \\
RKD \cite{park2019rkd} & & & 72.85$\pm$3.15 & 83.51$\pm$2.94 & 18.42$\pm$4.03 \\
CIRKD \cite{yang2022cirkd}& & & 74.04$\pm$3.03 & 85.06$\pm$2.92 & 14.43$\pm$3.53 \\
\rowcolor[HTML]{D4E6F1} \textbf{Ours(DPRD)} & & & \textbf{75.16$\pm$3.03} & \textbf{85.92$\pm$2.90} & \textbf{13.08$\pm$3.37} \\
\bottomrule
\end{tabular}
\end{table}

\begin{table}[!t]
\centering
\caption{Comparison of distillation methods on AMOS (CT) 2022.}
\label{tab:amos}
\begin{tabular}{l@{\hspace{-10pt}}cc@{\hspace{-1pt}}ccc}
\toprule
Method & Params(M) & FLOPs(G) & Dice(\%) $\uparrow$ & NSD(2mm) $\uparrow$ & HD95(mm) $\downarrow$ \\ 
\midrule
Teacher-MedNeXt &12.11 &  1360 & 85.37$\pm$2.03 & 84.93$\pm$2.14 & 14.52$\pm$2.12 \\
Baseline-MobileUNetV3&0.63 &39.19 &  80.16$\pm$2.58 & 77.87$\pm$3.09 & 18.74$\pm$1.69 \\
\hline
Logits KD \cite{hinton2015distilling}   &\multirow{5}{*}{0.63} &\multirow{5}{*}{39.19}  &83.14$\pm$2.42  &81.80$\pm$2.74  & 15.60$\pm$1.60  \\
FitNet \cite{romero2015fitnetshintsdeepnets}   & & &81.80$\pm$2.74  & 80.01$\pm$2.68  &  15.02$\pm$1.89  \\
RKD \cite{park2019rkd}& & & 83.49$\pm$2.29 &82.77$\pm$2.42  & 17.15$\pm$2.43 \\
CIRKD \cite{yang2022cirkd}& & & 82.09$\pm$2.54 & 80.13$\pm$2.67 & 14.75$\pm$1.46 \\
\rowcolor[HTML]{D4E6F1} \textbf{Ours(DPRD)} & & &\textbf{85.46$\pm$2.13} & \textbf{85.29$\pm$2.14} & \textbf{11.72$\pm$2.11} \\
\bottomrule
\end{tabular}
\end{table}

\textbf{Quantitative Results}
Tables~\ref{tab:isles}--\ref{tab:amos} summarize comparisons on ISLES 2022 and AMOS (CT) against standard distillation baselines.
On ISLES 2022, DPRD achieves the best Dice (75.16\%) and improves the PlainConvUNet student baseline (72.92\% Dice, 17.21 mm HD95) with the best surface scores among all methods (85.92\% NSD, 13.08\,mm HD95), the latter even surpassing the MedNeXt teacher (14.12\,mm HD95).
On AMOS, DPRD delivers the best overall performance (85.46\% Dice, 85.29\% NSD, 11.72\,mm HD95), outperforming all distillation baselines and slightly exceeding the MedNeXt teacher on Dice (85.37\%) while reducing boundary error (HD95 14.52$\rightarrow$11.72\,mm).
Overall, DPRD consistently strengthens compact students, with particularly pronounced gains on boundary-sensitive metrics.

\textbf{Qualitative Results} 
Fig. \ref{fig:isles_attention} visualizes encoder-stage responses on ISLES, where the late-stage student shows stronger focus on lesion regions than the early-stage student and better agreement with the ground-truth annotations. Fig. \ref{fig:amos_qual_case} further compares AMOS segmentations across KD baselines; DPRD produces masks more consistent with the ground truth, particularly around the liver, postcava, and gallbladder boundaries.

\begin{table}[htbp]
\centering
\caption{DPRD ablation study on AMOS 2022 (MobileUNetV3)}
\label{tab:dprd_ablation}

\setlength{\tabcolsep}{4.5pt}   
\renewcommand{\arraystretch}{1.1}

\begin{tabular}{lccc ccc}
\toprule
Method
& RAFM
& Traj.
& Dist.
& Dice (\%) $\uparrow$
& NSD (2mm) $\uparrow$
& HD95 (mm) $\downarrow$ \\
\midrule

Baseline
& \nmark & \nmark & \nmark
& 80.16$\pm$2.58
& 77.87$\pm$3.09
& 18.74$\pm$1.69 \\

\midrule

RAFM-off
& \nmark & \cmark & \cmark
& 81.68$\pm$2.67
& 79.72$\pm$2.84
& 21.65$\pm$2.74 \\

Distance-off
& \cmark & \cmark & \nmark
& 82.98$\pm$2.45
& 82.02$\pm$2.48
& 17.07$\pm$2.23 \\

Trajectory-off
& \cmark & \nmark & \cmark
& 83.04$\pm$2.39
& 82.21$\pm$2.42
& 15.03$\pm$2.03 \\

\midrule

\rowcolor[HTML]{D4E6F1}
\textbf{DPRD}
& \cmark & \cmark & \cmark
& \textbf{85.46$\pm$2.13}
& \textbf{85.29$\pm$2.14}
& \textbf{11.72$\pm$2.11} \\

\bottomrule
\end{tabular}
\end{table}

\textbf{Ablation Study}
Table \ref{tab:dprd_ablation} shows the contribution of each DPRD component. Removing RAFM decreases Dice from 85.46\% to 81.68\% and increases HD95 to 21.65 mm, indicating that background-dominated features can dilute the relational distillation signal. Removing either trajectory mapping or distance regularization also degrades performance, with HD95 increasing to 15.03 mm and 17.07 mm, respectively. The full \emph{DPRD} framework achieves the best HD95 of 11.72 mm, suggesting that directional trajectory reconstruction and relative distance consistency provide complementary supervision for accurate anatomical surface segmentation.

\section{Conclusion}

In this paper, we propose DPRD, an efficient distillation framework for robust 3D medical image segmentation. By combining ROI-aware feature masking with displacement-preserving relational alignment, DPRD improves compact students over feature-based and relational baselines on ISLES and AMOS, while achieving performance comparable to the MedNeXt teacher with substantially lower inference cost. A current limitation is that RAFM relies on ground-truth label masks during training, which may restrict its applicability in weakly-supervised or label-scarce settings; future work will explore teacher-predicted ROIs or uncertainty-guided masks to relax this requirement.

\begin{credits}
\subsubsection{\ackname}
The authors gratefully acknowledge the computational resources provided by the Delta system at the National Center for Supercomputing Applications (NCSA) [award OAC 2005572] through ACCESS allocations CIS260331 and CIS250976. ACCESS, the Advanced Cyberinfrastructure Coordination Ecosystem: Services \& Support program, is supported by U.S. National Science Foundation (NSF) grants \#2138259, \#2138286, \#2138307, \#2137603, and \#2138296. This work was partially supported by NIH grant U01AG079847 and NSF Award No.~2412285.

\subsubsection{\discintname}
The authors have no competing interests to declare that are relevant to the content of this article.

\end{credits}

\bibliographystyle{splncs04}
\bibliography{Paper-6226}

@article{mezger2013navigation,
  title={Navigation in surgery},
  author={Mezger, Uli and Jendrewski, Claudia and Bartels, Michael},
  journal={Langenbeck's archives of surgery},
  volume={398},
  number={4},
  pages={501--514},
  year={2013},
  publisher={Springer}
}

@article{li2026comprehensive,
  title={A Comprehensive Survey of Interaction Techniques in 3D Scene Generation},
  author={Li, Yuqi and Meng, Siwei and Yang, Chuanguang and Feng, Weilun and Liu, Junming and An, Zhulin and Wang, Yikai and Tian, Yingli},
  journal={Authorea Preprints},
  year={2026},
  publisher={Authorea}
}

@inproceedings{zhu2025pathology, title={Pathology-Aware Prototype Evolution via LLM-Driven Semantic Disambiguation for Multicenter Diabetic Retinopathy Diagnosis}, author={Zhu, Chunzheng and Lin, Yangfang and Shao, Jialin and Lin, Jianxin and Wang, Yijun}, booktitle={Proceedings of the 33rd ACM International Conference on Multimedia}, pages={9196--9205}, year={2025} }

@article{jia2026unsupervised, title={Unsupervised causal prototypical networks for de-biased interpretable dermoscopy diagnosis}, author={Jia, Junhao and Wu, Yueyi and Chen, Huangwei and Jing, Haodong and Wang, Haishuai and Bu, Jiajun and Wu, Lei}, journal={arXiv preprint arXiv:2602.23752}, year={2026} }

@inproceedings{lan2025acam,
  title={ACAM-KD: adaptive and cooperative attention masking for knowledge distillation},
  author={Lan, Qizhen and Tian, Qing},
  booktitle={Proceedings of the IEEE/CVF International Conference on Computer Vision},
  pages={3957--3966},
  year={2025}
}

@article{lan2026performance,
  title={From Performance to Practice: Knowledge-Distilled Segmentator for On-Premises Clinical Workflows},
  author={Lan, Qizhen and Choi, Aaron and Ma, Jun and Wang, Bo and Zhao, Zhaogming and Jiang, Xiaoqian and Hsu, Yu-Chun},
  journal={arXiv preprint arXiv:2601.09191},
  year={2026}
}

@article{nitsch2019mrius,
  title={Automatic and efficient MRI-US segmentations for improving intraoperative image fusion in image-guided neurosurgery},
  author={Nitsch, J and Klein, J and Dammann, P and Wrede, K and Gembruch, O and Moltz, JH and Meine, H and Sure, U and Kikinis, R and Miller, D},
  journal={NeuroImage: Clinical},
  volume={22},
  pages={101766},
  year={2019},
  publisher={Elsevier}
}

@inproceedings{roy2023mednext,
  title={Mednext: transformer-driven scaling of convnets for medical image segmentation},
  author={Roy, Saikat and Koehler, Gregor and Ulrich, Constantin and Baumgartner, Michael and Petersen, Jens and Isensee, Fabian and Jaeger, Paul F and Maier-Hein, Klaus H},
  booktitle={International conference on medical image computing and computer-assisted intervention},
  pages={405--415},
  year={2023},
  organization={Springer}
}

@article{isensee2021nnunet,
  title={nnU-Net: a self-configuring method for deep learning-based biomedical image segmentation},
  author={Isensee, Fabian and Jaeger, Paul F and Kohl, Simon AA and Petersen, Jens and Maier-Hein, Klaus H},
  journal={Nature methods},
  volume={18},
  number={2},
  pages={203--211},
  year={2021},
  publisher={Nature Publishing Group US New York}
}

@article{guan2021domain,
  title={Domain adaptation for medical image analysis: a survey},
  author={Guan, Hao and Liu, Mingxia},
  journal={IEEE Transactions on Biomedical Engineering},
  volume={69},
  number={3},
  pages={1173--1185},
  year={2021},
  publisher={IEEE}
}

@article{litjens2017survey,
  title={A survey on deep learning in medical image analysis},
  author={Litjens, Geert and Kooi, Thijs and Bejnordi, Babak Ehteshami and Setio, Arnaud Arindra Adiyoso and Ciompi, Francesco and Ghafoorian, Mohsen and Van Der Laak, Jeroen Awm and Van Ginneken, Bram and S{\'a}nchez, Clara I},
  journal={Medical image analysis},
  volume={42},
  pages={60--88},
  year={2017},
  publisher={Elsevier}
}

@inproceedings{milletari2016vnet,
  title={V-net: Fully convolutional neural networks for volumetric medical image segmentation},
  author={Milletari, Fausto and Navab, Nassir and Ahmadi, Seyed-Ahmad},
  booktitle={2016 fourth international conference on 3D vision (3DV)},
  pages={565--571},
  year={2016},
  organization={Ieee}
}

@article{hinton2015distilling,
  title  = {Distilling the Knowledge in a Neural Network},
  author = {Hinton, Geoffrey and Vinyals, Oriol and Dean, Jeff},
  journal= {arXiv preprint arXiv:1503.02531},
  year   = {2015}
}

@inproceedings{park2019rkd,
  title     = {Relational Knowledge Distillation},
  author    = {Park, Wonpyo and Kim, Dongju and Lu, Yan and Cho, Minsu},
  booktitle = {Proceedings of the IEEE/CVF Conference on Computer Vision and Pattern Recognition (CVPR)},
  pages     = {3967--3976},
  year      = {2019}
}

@inproceedings{yang2022cirkd,
  title     = {Cross-Image Relational Knowledge Distillation for Semantic Segmentation},
  author    = {Yang, Chuanguang and Zhou, Helong and An, Zhulin and Jiang, Xue and Xu, Yongjun and Zhang, Qian},
  booktitle = {Proceedings of the IEEE/CVF Conference on Computer Vision and Pattern Recognition (CVPR)},
  pages     = {12319--12328},
  year      = {2022}
}

@inproceedings{liu2019skdseg,
  title     = {Structured Knowledge Distillation for Semantic Segmentation},
  author    = {Liu, Yifan and Chen, Ke and Liu, Chris and Qin, Zengchang and Luo, Zhenbo and Wang, Jingdong},
  booktitle = {Proceedings of the IEEE/CVF Conference on Computer Vision and Pattern Recognition (CVPR)},
  pages     = {2604--2613},
  year      = {2019}
}

@inproceedings{shu2021cwd,
  title     = {Channel-Wise Knowledge Distillation for Dense Prediction},
  author    = {Shu, Changyong and Liu, Yifan and Gao, Jianfei and Yan, Zheng and Shen, Chunhua},
  booktitle = {Proceedings of the IEEE/CVF International Conference on Computer Vision (ICCV)},
  pages     = {5311--5320},
  year      = {2021}
}

@article{lee2022uxnet,
  title   = {3D UX-Net: A Large Kernel Volumetric ConvNet Modernizing Hierarchical Transformer for Medical Image Segmentation},
  author  = {Lee, H. H. and Bao, S. and Huo, Y. and Landman, B. A.},
  journal = {arXiv preprint arXiv:2209.15076},
  year    = {2022}
}

@inproceedings{hatamizadeh2022swinunetr,
  title={Swin unetr: Swin transformers for semantic segmentation of brain tumors in mri images},
  author={Hatamizadeh, Ali and Nath, Vishwesh and Tang, Yucheng and Yang, Dong and Roth, Holger R and Xu, Daguang},
  booktitle={International MICCAI brainlesion workshop},
  pages={272--284},
  year={2021},
  organization={Springer}
}

@article{dou2020_ummkd,
  title={Unpaired multi-modal segmentation via knowledge distillation},
  author={Dou, Qi and Liu, Quande and Heng, Pheng Ann and Glocker, Ben},
  journal={IEEE transactions on medical imaging},
  volume={39},
  number={7},
  pages={2415--2425},
  year={2020},
  publisher={IEEE}
}

@inproceedings{tung2019spkd,
  title={Similarity-preserving knowledge distillation},
  author={Tung, Frederick and Mori, Greg},
  booktitle={Proceedings of the IEEE/CVF international conference on computer vision},
  pages={1365--1374},
  year={2019}
}

@article{tian2020crd,
  title={Contrastive representation distillation},
  author={Tian, Yonglong and Krishnan, Dilip and Isola, Phillip},
  journal={arXiv preprint arXiv:1910.10699},
  year={2019}
}

@article{noothout2022_jmi_kdensembles,
  title={Knowledge distillation with ensembles of convolutional neural networks for medical image segmentation},
  author={Noothout, Julia MH and Lessmann, Nikolas and Van Eede, Matthijs C and Van Harten, Louis D and Sogancioglu, Ecem and Heslinga, Friso G and Veta, Mitko and Van Ginneken, Bram and I{\v{s}}gum, Ivana},
  journal={Journal of Medical Imaging},
  volume={9},
  number={5},
  pages={052407--052407},
  year={2022},
  publisher={Society of Photo-Optical Instrumentation Engineers}
}

@misc{romero2015fitnetshintsdeepnets,
      title={FitNets: Hints for Thin Deep Nets}, 
      author={Adriana Romero and Nicolas Ballas and Samira Ebrahimi Kahou and Antoine Chassang and Carlo Gatta and Yoshua Bengio},
      year={2015},
      eprint={1412.6550},
      archivePrefix={arXiv},
      primaryClass={cs.LG},
      url={https://arxiv.org/abs/1412.6550}, 
}

@inproceedings{he2019_kadseg,
  title={Knowledge adaptation for efficient semantic segmentation},
  author={He, Tong and Shen, Chunhua and Tian, Zhi and Gong, Dong and Sun, Changming and Yan, Youliang},
  booktitle={Proceedings of the IEEE/CVF conference on computer vision and pattern recognition},
  pages={578--587},
  year={2019}
}

@inproceedings{sudre2017_gdl,
  title={Generalised dice overlap as a deep learning loss function for highly unbalanced segmentations},
  author={Sudre, Carole H and Li, Wenqi and Vercauteren, Tom and Ourselin, Sebastien and Jorge Cardoso, M},
  booktitle={International Workshop on Deep Learning in Medical Image Analysis},
  pages={240--248},
  year={2017},
  organization={Springer}
}

@article{qin2021_tmi_kdseg,
  title={Efficient medical image segmentation based on knowledge distillation},
  author={Qin, Dian and Bu, Jia-Jun and Liu, Zhe and Shen, Xin and Zhou, Sheng and Gu, Jing-Jun and Wang, Zhi-Hua and Wu, Lei and Dai, Hui-Fen},
  journal={IEEE Transactions on Medical Imaging},
  volume={40},
  number={12},
  pages={3820--3831},
  year={2021},
  publisher={IEEE}
}

@article{petzsche2022isles,
  title={ISLES 2022: A multi-center magnetic resonance imaging stroke lesion segmentation dataset},
  author={Hernandez Petzsche, Moritz R and De La Rosa, Ezequiel and Hanning, Uta and Wiest, Roland and Valenzuela, Waldo and Reyes, Mauricio and Meyer, Maria and Liew, Sook-Lei and Kofler, Florian and Ezhov, Ivan and others},
  journal={Scientific data},
  volume={9},
  number={1},
  pages={762},
  year={2022},
  publisher={Nature Publishing Group UK London}
}

@article{ji2022amos,
  title={Amos: A large-scale abdominal multi-organ benchmark for versatile medical image segmentation},
  author={Ji, Yuanfeng and Bai, Haotian and Ge, Chongjian and Yang, Jie and Zhu, Ye and Zhang, Ruimao and Li, Zhen and Zhanng, Lingyan and Ma, Wanling and Wan, Xiang and others},
  journal={Advances in neural information processing systems},
  volume={35},
  pages={36722--36732},
  year={2022}
}

@inproceedings{ding2026dfm,
title     = {Difference Feature Map Distillation: Transferring Inter-Sample Relational Knowledge Towards Efficient Transformer-Based Tracking},
author    = {Ding, Zhicheng and Chu, Xinyu and Tian, Qing},
booktitle = {Proceedings of the 2026 IEEE Intelligent Vehicles Symposium (IV)},
year      = {2026}
}

\end{document}